# Logic in the Lab

Rineke Verbrugge
Institute of Artificial Intelligence
University of Groningen
P.O. Box 407
9700 AK Groningen,The Netherlands
L.C.Verbrugge@rug.nl

**Categories and Subject Descriptors**

F.4.1 [**Mathematical Logic**]: Modal Logic

**General Terms**

Theory

## 1. THEORY OF MIND

As humans, we live in a remarkably complex social environment. One cognitive tool which helps us manage all this complexity is our *theory of mind*, the ability to reason about the mental states of others. By deducing what other people want, feel and think, we can understand their actions, predict how our actions will influence them, and decide how we should behave to be successful. Theory of mind is the cognitive capacity to understand and predict external behavior of others and oneself by attributing internal mental states, such as knowledge, beliefs, and intentions [17]. This is thought to be the pinnacle of social cognition. A heated debate is going: Do very smart animals, such as chimpanzees and ravens, have any theory of mind? [3, 19].

Especially important in intelligent interaction is higher-order theory of mind, an agent's ability to model recursively mental states of other agents, including the other's model of the first agent's mental state, and so forth. More precisely, zero-order theory of mind concerns world facts, whereas $k+1$-order reasoning models $k$-order reasoning of the other agent or oneself. For example, "Bob knows that Alice knows that he wrote a novel under pseudonym" ($K_{Bob}K_{Alice}p$) is a second-order attribution. It is commonly accepted that animals other than human beings do *not* use second- and higher-order theory of mind.

Several formal theories well-known to the TARK audience are suited to represent higher-order theory of mind in intelligent interaction, for example, epistemic logic, dynamic epistemic logic, and epistemic game theory [15, 6, 22, 16]. However, in epistemic logic, unlimited rationality is usually taken for granted. Agents are assumed to be *logically omniscient*: they know all logical truths. The epistemic language allows reasoning on any modal depth and presupposes that agents can immediately decide whether a formula like $K_{Ann}\neg K_{Bob}K_{Ann}K_{Carol}\neg K_{Ann}\ w_{Ann}$ is true in a given possible world. This is clearly not the case for all people [23]. Similarly, people often do not act according to the game-theoretic assumption of common knowledge of rationality [4]. In particular, several researchers have found that both children and adults have difficulties when applying second-order theory of mind in game situations [10, 7]. But how do people really reason about others' mental states?

## 2. EXPERIMENTS

In our lab, we have performed several experiments with subjects applying second-order theory of mind in simple dynamic games. It turned out that we could facilitate their correct and fast performance a lot, for example, by providing step-wise training, by introducing a visual presentation that is easy to understand, and by prompting subjects to think about what their opponent would do [12, 13]. With the help of these cues, the subjects made the best possible decision in more than 90% of the game items. From what the subjects told us, however, we got the impression that even if they made the correct decisions, they did not reason exactly according to the game theory textbook. By a follow-up experiment with an eye-tracker, we concluded that indeed, most experimental subjects did not apply backward induction from the start, but tried to get by with forward reasoning as much as possible [14].

Formal methods are very useful for designing experiments and interpreting the results. As an example, Stenning and Van Lambalgen [18] provide an interesting analysis of the difficulties that autistic children have in ascribing false beliefs to another person, if they themselves know the true facts. As another example, one can investigate the computational complexity of the tasks that experimental subjects have been set [11]. Currently, we are investigating the complexity of several instances of backward induction and comparing them with subjects' behavior in terms of reaction times, decisions, and eye movements.

## 3. COGNITIVE MODELS

In order to understand how people really reason and solve problems, it has proven fruitful in cognitive psychology to use computational cognitive models implemented in a cognitive architecture such as ACT-R, which has been validated in hundreds of experiments [1]. It is also possible to use such computational models when investigating how people reason about other people's knowledge, beliefs and plans. One way to do this is to make an ACT-R model in which different reasoning strategies, such as backward reasoning and forward reasoning, 'compete' with one another and the model learns by experience which reasoning strategy efficiently provides effective decisions [9, 8]. The main advantage of using computational cognitive models is that one can formulate very precise predictions and see whether the simulations match results of new experiments in the lab.

This is just what we did in the case of the controversy about smart birds: Elske van der Vaart constructed a computational cognitive model of birds' smart social behavior. It turned out that this 'virtual bird', equipped with sophisticated memory based on the theory behind ACT-R [2], and reacting to the stress of being observed, performed similarly to the real birds in several experiments [20, 21]. In the literature, the birds' behavior is often thought to exemplify a form of perspective-taking: "I want to prevent that the other bird knows where I've hidden my worms" [5]. We made some precise predictions that can help settle the disputes between 'theory of mind' versus 'simple behavioral rules', and that are currently being investigated in the lab.

## Acknowledgments

I would like to thank the Netherlands Organization for Scientific Research (NWO) for Vici grant NWO 227-80-001, *Cognitive systems in interaction: Logical and computational models of higher-order social cognition.*

## 4. REFERENCES

[1] J. Anderson. *How Can the Human Mind Occur in the Physical Universe?* Oxford University Press, New York (NY), 2007.

[2] J. R. Anderson and L. J. Schooler. Reflections of the environment in memory. *Psychological Science*, 2(6):396–408, 1991.

[3] J. Call and M. Tomasello. Does the chimpanzee have a theory of mind? 30 years later. *Trends in Cognitive Sciences*, 12:187–192, 2008.

[4] C. Camerer. *Behavioral Game Theory: Experiments in Strategic Interaction*. Princeton University Press, Princeton (NJ), 2003.

[5] J. Dally, N. Emery, and N. Clayton. Food-caching western scrub-jays keep track of who was watching when. *Science*, 312:1662–1665, 2006.

[6] R. Fagin, J. Halpern, Y. Moses, and M. Vardi. *Reasoning about Knowledge*. MIT Press, Cambridge, MA, 1995. Second edition 2003.

[7] L. Flobbe, R. Verbrugge, P. Hendriks, and I. Krämer. Children's application of theory of mind in reasoning and language. *Journal of Logic, Language and Information*, 17:417–442, 2008. Special issue on formal models for real people, edited by M. Counihan.

[8] S. Ghosh and B. Meijering. On combining cognitive and formal modeling: A case study involving strategic reasoning. In J. van Eijck and R. Verbrugge, editors, *Proceedings of the Workshop on Reasoning About Other Minds (RAOM 2011)*, volume 751, pages 79–92. CEUR Workshop Proceedings, 2011.

[9] S. Ghosh, B. Meijering, and R. Verbrugge. Logic meets cognition: Empirical reasoning in games. In *Proceedings of the 3rd International Workshop on Logics for Resource Bounded Agents (LRBA 2010), in 3rd Multi-Agent Logics, Languages, and Organisations Federated Workshops, MALLOW'10, CEUR Workshop Proceedings*, volume 627, pages 15–34, 2010.

[10] T. Hedden and J. Zhang. What do you think I think you think? Strategic reasoning in matrix games. *Cognition*, 85:1–36, 2002.

[11] A. Isaac, J. Szymanik, and R. Verbrugge. Logic and complexity in cognitive science. In *Logical and Informational Dynamics: Johan van Benthem*, Trends in Logic: Outstanding Contributions, Berlin, 2013, to appear. Springer.

[12] B. Meijering, L. v. Maanen, H. v. Rijn, and R. Verbrugge. The facilitative effect of context on second-order social reasoning. In *Proceedings of the 32nd Annual Meeting of the Cognitive Science Society*, pages 1423–1428, Philadelphia, PA, 2010. Cognitive Science Society.

[13] B. Meijering, H. v. Rijn, N. Taatgen, and R. Verbrugge. I do know what you think I think: Second-order theory of mind in strategic games is not that difficult. In *Proceedings of the 33nd Annual Meeting of the Cognitive Science Society*, pages 2486–2491, Austin, TX, 2011. Cognitive Science Society.

[14] B. Meijering, H. van Rijn, N. Taatgen, and R. Verbrugge. What eye movements can tell about theory of mind in a strategic game. *PLoS ONE*, In press.

[15] J.-J. C. Meyer and W. van der Hoek. *Epistemic Logic for AI and Theoretical Computer Science*. Cambridge University Press, Cambridge, 1995.

[16] A. Perea. *Epistemic Game Theory*. Cambridge University Press, Cambridge, 2012.

[17] D. Premack and G. Woodruff. Does the chimpanzee have a theory of mind? *Behavioral and Brain Sciences*, 4:515–526, 1978.

[18] K. Stenning and M. van Lambalgen. *Human Reasoning and Cognitive Science*. MIT Press, Cambridge (MA), 2008.

[19] E. van der Vaart and C. Hemelrijk. 'Theory of mind' in animals: Ways to make progress. *Synthese*, pages 1–20, 2012.

[20] E. van der Vaart, R. Verbrugge, and C. Hemelrijk. Corvid caching: Insights from a cognitive model. *Journal of Experimental Psychology: Animal Behavior Processes*, 37(3):330, 2011.

[21] E. van der Vaart, R. Verbrugge, and C. Hemelrijk. Corvid re-caching without 'theory of mind': A model. *PloS ONE*, 7(3):e32904, 2012.

[22] H. van Ditmarsch, W. van der Hoek, and B. Kooi. *Dynamic Epistemic Logic*, volume 337 of *Synthese Library Series*. Springer Verlag, Berlin, 2007.

[23] R. Verbrugge. Logic and social cognition: The facts matter, and so do computational models. *Journal of Philosophical Logic*, 38(6):649–680, 2009.